\documentclass[letterpaper]{article} 
\usepackage{aaai24}  
\usepackage{times}  
\usepackage{helvet}  
\usepackage{courier}  
\usepackage[hyphens]{url}  
\usepackage{array}
\usepackage{floatrow}
\usepackage{lipsum} 
\usepackage{multicol} 
\usepackage{subcaption}
\newcolumntype{L}[1]{>{\raggedright\let\newline\\\arraybackslash\hspace{0pt}}m{#1}}
\newcolumntype{C}[1]{>{\centering\let\newline\\\arraybackslash\hspace{0pt}}m{#1}}
\newcolumntype{M}[1]{>{\centering\arraybackslash}m{#1}}

\usepackage{graphicx} 
\usepackage{natbib}  
\usepackage{caption} 
\usepackage{algorithm}
\usepackage{algpseudocode} 
\renewcommand{\algorithmicrequire}{\textbf{Input:}} 
\usepackage{footnote}
\usepackage{pifont}
\makesavenoteenv{algorithm}
\usepackage{afterpage}

\usepackage{amsmath}
\usepackage{amssymb}
\usepackage{amsbsy}
\usepackage{mathtools} 
\usepackage{amsthm} 
\usepackage{booktabs} 
\usepackage{multirow} 
\usepackage{adjustbox}
\usepackage{arydshln}
\usepackage{dsfont}

\usepackage{newfloat}
\usepackage{listings}
\DeclareCaptionStyle{ruled}{labelfont=normalfont,labelsep=colon,strut=off} 
\floatstyle{ruled}
\newfloat{listing}{tb}{lst}{}
\floatname{listing}{Listing}
\title{REPrune: Channel Pruning via Kernel Representative Selection}
\author {
    Mincheol Park\textsuperscript{\rm 1,\rm 3},
    Dongjin Kim\textsuperscript{\rm 2,\rm 3},
    Cheonjun Park\textsuperscript{\rm 1},
    Yuna Park\textsuperscript{\rm 3}, \\
    Gyeong Eun Gong\textsuperscript{\rm 4},
    Won Woo Ro\textsuperscript{\rm 1},
    and Suhyun Kim\textsuperscript{\rm 3}\equalcontrib
}
\affiliations {
    \textsuperscript{\rm 1}Yonsei University, Republic of Korea \\
    \textsuperscript{\rm 2}Korea University, Republic of Korea \\
    \textsuperscript{\rm 3}Korea Institute of Science and Technology, Republic of Korea \\
    \textsuperscript{\rm 4}Hyundai MOBIS, Republic of Korea \\
    \{mincheol.park, cheonjun.park, wro\}@yonsei.ac.kr, npclinic3@korea.ac.kr \\
    \{qkrdbsdk0427, dr.suhyun\_kim\}@gmail.com, gegong@mobis.co.kr
}

\begin{document}

\maketitle

\begin{abstract}
Channel pruning is widely accepted to accelerate modern convolutional neural networks (CNNs). The resulting pruned model benefits from its immediate deployment on general-purpose software and hardware resources. However, its large pruning granularity, specifically at the unit of a convolution filter, often leads to undesirable accuracy drops due to the inflexibility of deciding how and where to introduce sparsity to the CNNs. In this paper, we propose REPrune, a novel channel pruning technique that emulates kernel pruning, fully exploiting the finer but structured granularity. REPrune identifies similar kernels within each channel using agglomerative clustering. Then, it selects filters that maximize the incorporation of kernel representatives while optimizing the maximum cluster coverage problem. By integrating with a simultaneous training-pruning paradigm, REPrune promotes efficient, progressive pruning throughout training CNNs, avoiding the conventional train-prune-finetune sequence. Experimental results highlight that REPrune performs better in computer vision tasks than existing methods, effectively achieving a balance between acceleration ratio and performance retention.

\end{abstract}

\section{Introduction}
The growing utilization of convolutional layers in modern CNN architectures presents substantial challenges for deploying these models on low-power devices.
As convolution layers account for over 90\% of the total computation volume, CNNs necessitate frequent memory accesses to handle weights and feature maps, resulting in an increase in hardware energy consumption~\cite{chen2016eyeriss}.
To address these challenges, network pruning is being widely explored as a viable solution for model compression, which aims to deploy CNNs on energy-constrained devices.

Channel pruning~\cite{you2019gate,chin2020towards,sui2021chip,hou2022chex} in network pruning techniques stands out as a practical approach. It discards redundant channels—each corresponding to a convolution filter—in the original CNNs, leading to a dense, narrow, and memory-efficient architecture. Consequently, this pruned sub-model is readily deployable on general-purpose hardware, circumventing the need for extra software optimizations~\cite{song2016deepcompression}.
Moreover, channel pruning enhances pruning efficiency by adopting a concurrent training-pruning pipeline.
This method entails the incremental pruning of low-ranked channels in training time.

\begin{figure}[t!]
    \centering
        \includegraphics[width=1\linewidth]{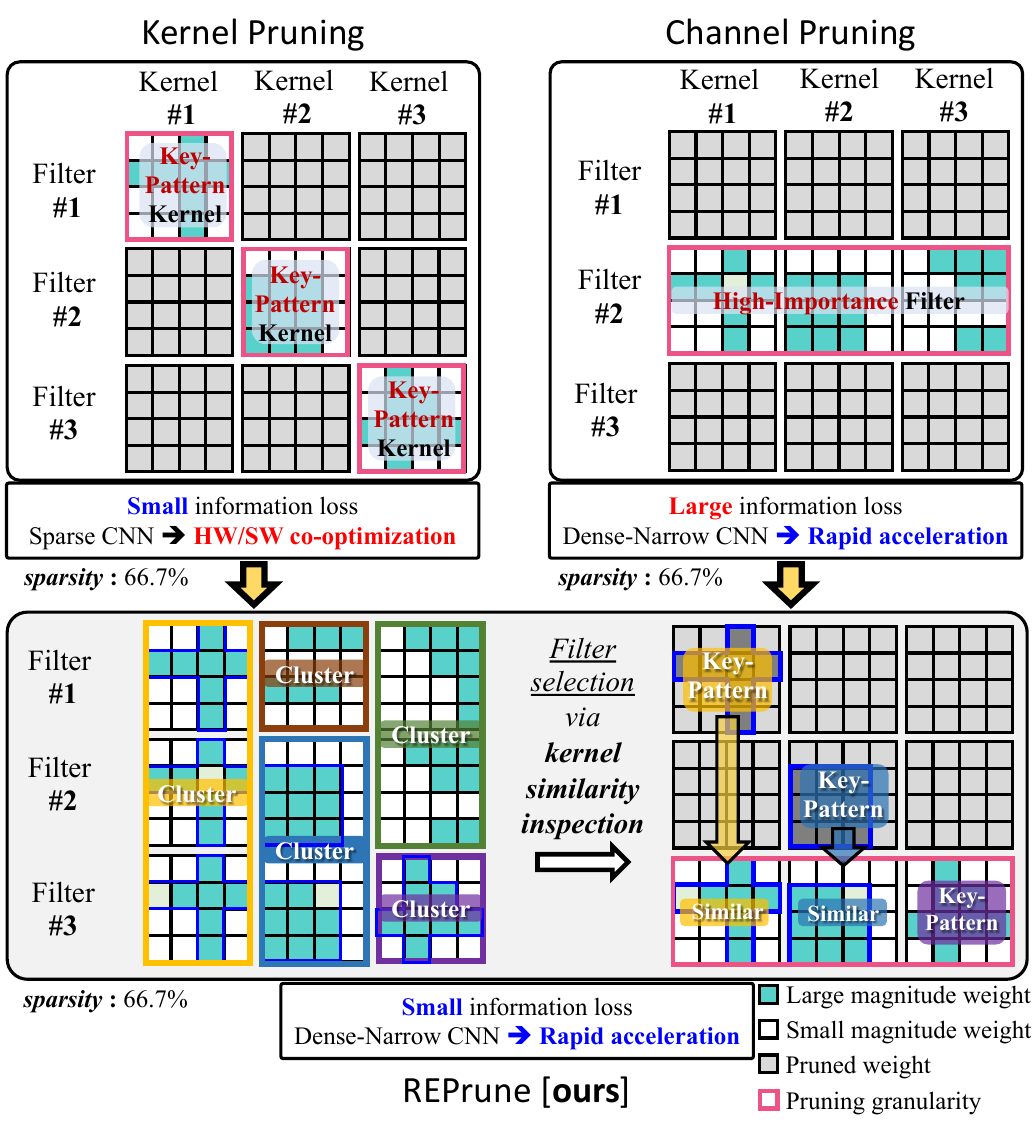}
    \caption{To accelerate CNN and minimize its information loss simultaneously, REPrune intends to select filters associated with patterned key kernels targeted by kernel pruning.}
    \label{fig_1}
\end{figure}

However, pruning a channel (a filter unit), which exhibits larger granularity, encounters greater challenges in preserving accuracy than the method with smaller granularity~\cite{park2023balanced}. This larger granularity often results in undesirable accuracy drops due to the inflexibility in how and where to introduce sparsity into the original model~\cite{zhong2022revisit}. Hence, an approach with a bit of fine-grained pruning granularity, such as kernel pruning~\cite{ma2020pconv,niu2020patdnn} illustrated in Fig.~\ref{fig_1}, has been explored within the structured pruning domain to overcome these limitations.

This paper focuses on such a kernel, a slice of the 3D tensor of a convolution filter, offering a more fine-grained granularity than an entire filter.
Unlike filters, whose dimensions vary a lot according to the layer depth, kernels in modern CNN architectures have exact dimensions, typically $3\times 3$ or $1\times 1$, across all layers.
This consistency allows for the fair application of similarity decision criteria.
However, naively pruning non-critical kernels~\cite{niu2020patdnn,zhong2022revisit} makes CNNs no longer dense. These sparse models require the support of code regeneration or load-balancing techniques to ensure parallel processing so as to accelerate inference~\cite{niu2020patdnn,park2023balanced}.

Our goal is to develop a channel pruning technique that implies fine-grained kernel inspection.
We intend to identify similar kernels and leave one between them.
Fig.\ref{fig_1} illustrates an example of critical diagonal kernels being selected via kernel pruning.
These kernels may also display high similarity with others in each channel, resulting in clusters~\cite{li2019exploiting} as shown in Fig.\ref{fig_1}.
By fully exploiting the clusters, we can obtain the effect of leaving diagonal kernels by selecting the bottom filter, even if it is not kernel pruning.
It is worth noting that this selection is not always feasible, but this way is essential to leave the best kernel to produce a densely narrow CNN immediately.
In this manner, we can finally represent channel pruning emulating kernel pruning; this is the first attempt to the best of our knowledge.

\begin{figure*}[!t]
    \centering
        \includegraphics[width=1.0\linewidth]{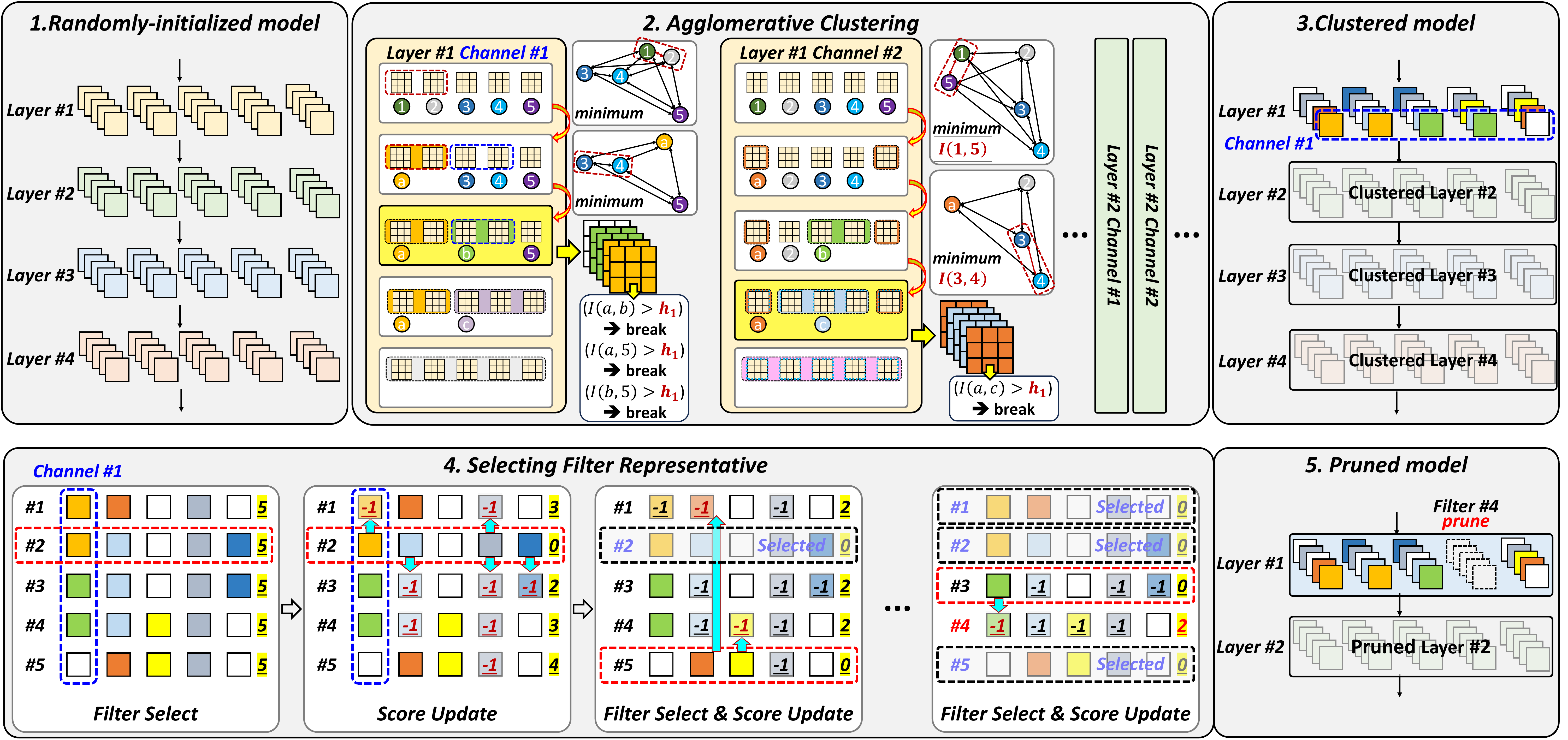}
    \caption{An overview of the REPrune methodology for identifying redundant kernels and selecting filters. Every channel performs agglomerative clustering on its corresponding kernel set in each layer. Once clusters are formed in accordance with the target channel sparsity, our proposed solver for the MCP starts its greedy filter selection until the target number of channels is satisfied. This solver selects a filter that includes a representative kernel from each grouped cluster per channel.}
    \label{fig_reprune_algorithm}
\end{figure*}

This paper proposes a novel channel pruning technique entitled ``REPrune.'' REPrune selects filters incorporating the most representative kernels, as depicted in Fig.~\ref{fig_reprune_algorithm}. 
REPrune starts with agglomerative clustering on a kernel set that generates features of the same input channel.
Agglomerative clustering provides a consistent hierarchical linkage.
By the linkage sequence, we can get an extent of similarity among kernels belonging to the same cluster.
This consistency allows for introducing linkage cut-offs to break the clustering process when the similarity exceeds a certain degree while the clustering process intertwines with a progressive pruning framework~\cite{hou2022chex}.
More specifically, these cut-offs act as linkage thresholds, determined by the layer-wise target channel sparsity inherent in this framework to obtain per-channel clusters.
Then, REPrune selects a filter that maximizes the coverage for kernel representatives from all clusters in each layer while optimizing our proposed maximum coverage problem.
Our comprehensive evaluation demonstrates that REPrune performs better in computer vision tasks than previous methods.

Our contributions can be summarized as three-fold:
\begin{itemize}
    \item We present REPrune, a channel pruning method that identifies similar kernels based on clustering and selects filters covering their representatives to achieve immediate model acceleration on general-purpose hardware.
    \item We embed REPrune within a concurrent training-pruning framework, enabling efficiently pruned model derivation in just one training phase, circumventing the traditional train-prune-finetune procedure.
    \item REPrune also emulates kernel pruning attributes, achieving a high acceleration ratio with well-maintaining performance in computer vision tasks.
\end{itemize}

\section{Related Work}

\paragraph{Channel pruning}
Numerous techniques have been developed to retain a minimal, influential set of filters by identifying essential input channels during the training phase of CNNs. Within this domain, training-based methods~\cite{liu2017learning,he2017channel,you2019gate,li2020group} strive to identify critical filters by imposing LASSO or group LASSO penalties on the channels. However, this regularization falls short of enforcing the exact zeroing of filters. To induce complete zeros for filters, certain studies~\cite{ye2018rethinking,lin2019towards} resort to a proxy problem with ISTA~\cite{beck2009fast}. Other importance-based approaches try to find optimal channels through ranking~\cite{lin2020hrank}, filter norm~\cite{he2018soft,he2019filter}, or mutual independence~\cite{sui2021chip}, necessitating an additional fine-tuning process. To fully automate the pruning procedure, sampling-based methods~\cite{kang2020operation,gao2020discrete} approximate the channel removal process. These methods automate the selection process of channels using differentiable parameters. In contrast, REPrune, without relying on differentiable parameters, simplifies the automated generation of pruned models through a progressive channel pruning paradigm~\cite{hou2022chex}.

\paragraph{Clustering-based pruning}
Traditional techniques~\cite{Duggal2019CUPCP,lee2020filter} exploit differences in filter distributions across layers~\cite{he2019filter} to identify similarities between channels. These methods group similar filters or representations, mapping them to a latent space, and retain only the essential ones.
Recent studies~\cite{chang2022automatic,wu2022filter} during the training phase of CNNs cluster analogous feature maps to group the corresponding channels.
Their aim is to ascertain the optimal channel count for each layer, which aids in designing a pruned model.
REPrune uses clustering that is nothing too novel.
In other words, REPrune departs from the filter clustering but performs kernel clustering in each input channel to ultimately prune filters. This is a distinct aspect of REPrune.

\paragraph{Kernel pruning}
Most existing methods~\cite{ma2020pconv,niu2020patdnn} retain specific patterns within convolution kernels, subsequently discarding entire connections associated with the least essential kernels. To improve kernel pruning, previous studies~\cite{yu2017accelerating,zhang2022groupbasednp} have initiated the process by grouping filters based on their similarity.
A recent work~\cite{zhong2022revisit} combines the Lottery Ticket Hypothesis~\cite{frankle2018the} to group cohesive filters early on, then optimizes kernel selection through their specialized weight selection process.
However, such methodologies tend to produce sparse CNNs, which demand dedicated solutions~\cite{park2023balanced} for efficient execution. Diverging from this challenge, REPrune focuses on selecting filters that best represent critical kernels, circumventing the need for extra post-optimization.


\section{Methodology} \label{method}

\paragraph{Prerequisite} We denote the number of input and output channels of the $l$-th convolutional layer as $n_l$ and $n_{l+1}$, respectively, with $k_h \times k_w$ representing the kernel dimension of the $l$-th convolution filters. A filter set for the $l$-th layer are represented as $\mathcal{F}^{l} \in \mathbb{R}^{n_{l+1}\times n_l\times k_h \times k_w}$. We define the $j$-th set of kernels of $\mathcal{F}^l$ as $\mathcal{K}^l_j \in \mathbb{R}^{n_{l+1}\times k_h \times k_w}$, where $j$ refers to the index of an input channel. Each individual kernel is denoted by $\kappa^{l}_{i,j}\in\mathbb{R}^{k_h \times k_w}$, resulting in $\mathcal{K}^l_j=\{ \kappa^{l}_{i,j}: \forall i \in \mathcal{I} \}$. Lastly, we define $\mathcal{I}=\{ 1,...,n_{l+1} \}$ and $\mathcal{J}=\{ 1,...,n_{l} \}$.

\subsection{Preliminary: Agglomerative Clustering} \label{agglomerative_clutsering}
Agglomerative clustering acts on `bottom-up' hierarchical clustering by starting with singleton clusters (i.e., separated kernels) and then iteratively merging two clusters into one until only one cluster remains.

Let a \textit{cluster} set after the agglomerative clustering repeats $c$-th merging process for the $j$-th kernel set $\mathcal{K}^l_j$ as $\mathcal{AC}_c(\mathcal{K}^{l}_{j})$. This process yields a sequence of intermediate cluster sets $\mathcal{AC}_0(\mathcal{K}^{l}_{j}), ...,\mathcal{AC}_{n_{l+1}-1}(\mathcal{K}^l_j)$ where $\mathcal{AC}_0(\mathcal{K}^{l}_{j})=\{ \{ \kappa^{l}_{i,j} \}: \forall\kappa^{l}_{i,j}\in \mathcal{K}^{l}_{j} \}$ denotes the initial set of individual kernels, and $\mathcal{AC}_{n_{l+1}-1}(\mathcal{K}^l_j)=\{ \mathcal{K}^l_j \}$ is the root cluster that incorporates all kernels in $\mathcal{K}^l_j$. This clustering process can be typically encapsulated in the following recurrence relationship:
\begin{equation} \label{agg-recurrence}
    \mathcal{AC}_c(\mathcal{K}^{l}_{j}) = \mathcal{AC}_{c-1}(\mathcal{K}^{l}_{j})\setminus\{ A_c, B_c \}\cup\{ A_c \cup B_c \},
\end{equation}
where $A_c$ and $B_c$ denotes two arbitrary clusters from $\mathcal{AC}_{c-1}(\mathcal{K}^{l}_{j})$, which are the closest in the merging step $c$. The distance between two clusters can be measured using the single linkage, complete linkage, average linkage, and Ward's linkage distance~\cite{witten2002data}.

This paper exploits Ward's method~\cite{ward1963hierarchical}.
First, it prioritizes that kernels within two clusters are generally not too dispersed in merging clusters~\cite{murtagh2014ward}.
In other words, Ward's method can calculate the dissimilarity of all combinations of the two clusters at each merging step $c$.
Thus, at each merging step $c$, a pair of clusters to be grouped can be identified based on the maximum similarity value, which represents the smallest relative distance compared to other distances.
Ward's linkage distance defines $I(A_c, B_c)=\Delta(A_c\cup B_c)-\Delta(A_c)-\Delta(B_c)$ where $\Delta(\cdot)$ is the sum of squared error. Then, the distance between two clusters is defined as follows:
\begin{equation}\label{ward}
\begin{split}
I(A_c, B_c) & = \frac{\left | A_c \right | \left | B_c \right |}{\left | A_c \right | + \left |  B_c \right |} {\left \| \mathbf{m}_{A_c} - \mathbf{m}_{B_c} \right \|}_2.
\end{split}
\end{equation}
where $\mathbf{m}_{A_c}$ and $\mathbf{m}_{B_c}$ are the centroid of each cluster and $\left | \cdot \right |$ is the number of kernels in them. Hence, Ward's linkage distance starts at zero when every kernel is a singleton cluster and grows as we merge clusters hierarchically. 
Second, Ward’s method ensures this growth monotonically~\cite{milligan1979ultrametric}. Let $d(c;\mathcal{K}^l_j)$, the minimum of Ward's linkage distances, be the linkage objective function in the $c$-th linkage step: $d(c;\mathcal{K}^l_j) = \text{min}_{(A_c, B_c)\in\mathcal{AC}_{c-1}(\mathcal{K}^l_j)} I(A_c, B_c)$, s.t., $A_c\neq B_c$, $\forall c \in \{1,...,n_{l+1}-1\}$ and $d(0;\mathcal{K}^l_j) = 0$. Then, $d(c-1;\mathcal{K}^l_j) \leq d(c;\mathcal{K}^l_j)$. The non-decreasing property of Ward's linkage distance leads to a consistent linkage order, although the clustering repeats. By the monotonicity of Ward's method, agglomerative clustering can even perform until it reaches a cut-off height $h$ as follows:
\begin{equation} \label{cut-off}
    {\mathbf{1}}_c(h;\mathcal{K}^{l}_{j}) = 
    \begin{cases}
    \displaystyle \mathcal{AC}_c(\mathcal{K}^{l}_{j}), & h \geq d(c;\mathcal{K}^l_j), \\
                  \mathcal{AC}_{c-1}(\mathcal{K}^{l}_{j}), & \text{otherwise}, \\
    \end{cases}
\end{equation}
where $\mathbf{1}_c(\cdot)$ is a linkage control function to perform agglomerative clustering in $c$-th merging step.
The monotonicity of Ward's method allows for a direct mapping of Ward's linkage distance value to the height $h$, which can then be used as a distance value.
Therefore, Eq.\ref{cut-off} can be used to generate available clusters until $d(c;\mathcal{K}^l_j)$ does not exceed $h$ as the desired cut-off parameter.

The following section will introduce how we set $h$ as the linkage cut-off to determine per-channel clusters, and then REPrune makes the per-channel clusters through Eq.~\ref{cut-off}.

\subsection{Foundation of Clusters Per Channel}

This section introduces how to set a layer-specific linkage cut-off, termed $h_l$, and demonstrates the process to produce clusters per channel using $h_l$ in each $l$-th convolutional layer.
Instead of applying a uniform $h$ across all layers~\cite{Duggal2019CUPCP}, our strategy employs a unique $h_l$ to break the clustering process when necessary in each layer.
This ensures that the distances between newly formed clusters remain out of a threshold so as to preserve an acceptable degree of dispersion in each layer accordingly.

Our method begins with agglomerative clustering on each kernel set $\mathcal{K}^l_j$. Given the $l$-th layer's channel sparsity $s^l \in [0, 1]$\footnotemark[1], kernel sets continue to cluster until $\lceil (1-s^l)n_{l+1} \rceil$ clusters form\footnotemark[2]. In other words, agglomerative clustering needs to repeat $\tilde{n}_{l+1}$ (where $\tilde{n}_{l+1}=\lceil s^ln_{l+1} \rceil$) merger steps.
In this paper, we make the cluster set per channel as $\mathcal{AC}_{\tilde{n}_{l+1}}(\mathcal{K}^l_1)$ up to $\mathcal{AC}_{\tilde{n}_{l+1}}(\mathcal{K}^{l}_{n_l})$.
\footnotetext[1]{The process for obtaining $s^l$ will be outlined in detail within the overview of the entire pipeline in this section.}

At this merging step $\tilde{n}_{l+1}$, we collect Ward's linkage distances, denoted as $d(\tilde{n}_{l+1}; \mathcal{K}^l_1)$ through $d(\tilde{n}_{l+1}; \mathcal{K}^l_{n_l})$.
Note that kernel distributions can vary significantly across channels~\cite{li2019exploiting}.
This means some channels may generate distances that are too close, reflecting the cohesiveness of the kernels, while others may yield greater distances due to the diversity of the kernels.

The diversity of kernels can lead to variations in the number of clusters for each channel. Furthermore, a smaller cut-off height tends to increase this number of clusters, which indicates a high preservation of kernel representation. To both preserve the representation of each channel and ensure channel sparsity simultaneously, we set $h_l$ to the maximum value among the collected distances as follows:
\begin{equation} \label{acc_sse}
    h_l=\text{max}_{j\in\mathcal{J}} d(\tilde{n}_{l+1}; \mathcal{K}^l_j), \quad l \in \mathcal{L}.
\end{equation}
The linkage cut-off $h_l$ serves as a pivotal parameter in identifying cluster sets in each channel based on the channel sparsity $s^l$.
This linkage cut-off $h_l$ aids in choosing suitable cluster set $ \mathcal{AC}^*(\mathcal{K}^l_j)$ between $\mathcal{AC}_{\tilde{n}_{l+1}-1}(\mathcal{K}^l_j)$ and $\mathcal{AC}_{\tilde{n}_{l+1}}(\mathcal{K}^l_j)$, using the control function defined in Eq.~\ref{cut-off}:
\begin{equation} \label{per_channel_clusters}
    \mathcal{AC}^*(\mathcal{K}^l_j) = {\mathbf{1}}_{\tilde{n}_{l+1}}(h_l; \mathcal{K}^l_j), \quad  j \in \mathcal{J}, l \in \mathcal{L}. 
\end{equation}
Each cluster, derived from Eq.~\ref{per_channel_clusters}, indicates a group of similar kernels, any of which can act as a representative within its cluster. The following section provides insight into selecting a filter that optimally surrounds these representatives.

\subsection{Filter Selection via Maximum Cluster Coverage} \label{select_rep_filter}
This section aims to identify a subset of the filter set $\mathcal{F}^l$ to yield the maximum coverage for kernel representatives across all clusters within each $l$-th convolutional layer.

We frame this objective within the Maximum Coverage Problem (MCP).
In this formulation, each kernel in a filter is directly mapped to a distinct cluster.
Further, a \textit{coverage score}, either 0 or 1, is allocated to each kernel.
This score indicates whether the cluster corresponding to a given kernel has already been represented as filters are selected.
Therefore, our primary strategy for optimizing the MCP involves prioritizing the selection of filters that maximize the sum of these coverage scores.
This way, we approximate an entire representation of all clusters within the reserved filters.

To define our MCP, we introduce a set of clusters $U^l$ from the $l$-th convolutional layer.
This set, denoted as $U^l = \{ \mathcal{AC}^*(\mathcal{K}^{l}_{1});\cdots; \mathcal{AC}^*(\mathcal{K}^{l}_{n_l}) \}$, represents the clusters that need to be covered.
Given a filter set $\mathcal{F}^l$, each $j$-th kernel $\kappa^l_{i,j}$ of a filter corresponds to a cluster in the set $\mathcal{AC}^*(\mathcal{K}^{l}_{j})$, where $\forall i\in\mathcal{I}$ and $\forall j\in\mathcal{J}$.
This way, we can view each filter as a subset of $U^l$.
This perspective leads to the subsequent set cover problem, which aims to approximate $U^l$ using an optimal set $\tilde{\mathcal{F}}^l$ that contains only the necessary filters.

The objective of our proposed MCP can be cast as a minimization problem that involves a subset of filters, denoted as $\tilde{\mathcal{F}}^l\subset \mathcal{F}^l$, where $\tilde{\mathcal{F}}^l\in \mathbb{R}^{\lceil(1-s^l){n}_{l+1}\rceil \times n_l\times k_h\times k_w}$ represents the group of filters in the $l$-th layer that are not pruned:
\begin{equation} \label{mcp}
    \begin{gathered}
    \text{min} \left( |U^l|-\sum_{\mathcal{F}^l_r\in\tilde{\mathcal{F}}^l}\sum_{\kappa^l_{r,j}\in\mathcal{F}^l_r}\sum_{j\in\mathcal{J}}S(\kappa^l_{r,j}) \right), \\
    \text{s.t.}, S(\kappa^l_{r,j}) \in \{0, 1\}, \; | \tilde{\mathcal{F}}^l | = \lceil(1-s^l){n}_{l+1}\rceil,
    \end{gathered}
\end{equation}
where $\mathcal{F}^l_r$ is a selected filter, and $|U^l|=\sum^{n_l}_{j=1}|AC^*(\mathcal{K}^l_j)|$ is the sum of optimal coverage scores.
\textit{Each kernel in $\mathcal{F}^l_r$ initially has a coverage score, denoted by $S(\kappa^l_{r,j})$, of one. This score transitions to zero if the cluster from $\mathcal{AC}^*(\mathcal{K}^l_j)$, to which $\kappa^l_{r,j}$ maps, is already covered}.
\footnotetext[2]{The ceiling function is equal to $\lceil x \rceil=\text{min}\{ n\in\mathbb{Z}: n \geq x \}$.}

As depicted in Fig.~\ref{fig_reprune_algorithm}, we propose a greedy algorithm to optimize Eq.~\ref{mcp}.
This algorithm selects filters encompassing the maximum number of uncovered clusters as follows:
\begin{equation} \label{greedy}
    \begin{gathered}
    i = \text{argmax}_{i\in\mathcal{I}}\sum\nolimits_{j \in \mathcal{J}}S(\kappa^l_{i,j}), \quad \forall\kappa^l_{i,j}\in \mathcal{F}^l_i.
    \end{gathered}
\end{equation}
There may be several candidate filters, $\mathcal{F}^l_{i}, \mathcal{F}^l_{i'}, ..., \mathcal{F}^l_{i''}$ (where $i\neq i' \neq \cdots \neq i''$), that share the maximum, identical coverage scores.
Given that any filter among these candidates could be equally representative, we resort to random sampling to select a filter $\mathcal{F}^l_r$ from them.

Upon the selection and inclusion of a filter $\mathcal{F}^l_r$ into $\tilde{\mathcal{F}}^l$, the coverage scores of each kernel in the remaining filters from $\mathcal{F}^l$ are updated accordingly. 
This update continues in the repetition of the procedure specified in Eq.~\ref{greedy} until a total of $\lceil(1-s^l){n}_{l+1}\rceil$ filters have been selected.
This thorough process is detailed in Alg.\ref{REPrune}.

\begin{algorithm}[!t]
\caption{Channel Selection in REPrune}\label{REPrune}
\begin{algorithmic}[1]
\Require{Filter set $\mathcal{F}^l=\{\mathcal{K}^l_j:  j\in\mathcal{J}\}, \forall l\in \mathcal{L}$}
\Ensure{Non-pruned filter set $\tilde{\mathcal{F}}^l, \; \forall l\in \mathcal{L}$\footnotemark}
\renewcommand{\algorithmicrequire}{\textbf{Given:}} 
\Require{$\mathcal{AC}$; Target channel sparsity $s^l, \; \forall l\in \mathcal{L}$}
\For{layer $l \in \mathcal{L}$}
    \For{channel $j$ \textbf{from} $1$ \textbf{to} $n_{l}$}
        \State{$\mathcal{AC}_{c}(\mathcal{K}^l_j)$ until merging step $c$ becomes $\tilde{n}_{l+1}$}
        \State{Get $d(\tilde{n}_{l+1}; \mathcal{K}^l_j)$ and then obtain a cut-off $h_l$}
        \State{Set clusters per channel $\mathcal{AC}^*(\mathcal{K}^l_j)$ \algorithmiccomment{Eq.~\ref{per_channel_clusters}}}
    \EndFor
    
    \State{Define the optimal cluster coverage $U^l$}
    \State{Initialize $\tilde{\mathcal{F}}^l$ as an empty queue}
    \While{$|\tilde{\mathcal{F}}^l| < \lceil(1-s^l){n}_{l+1}\rceil$}
        \State{Select candidate filters in $\mathcal{F}^l$ \algorithmiccomment{Eq.~\ref{greedy}} }
        \State{Sample $\mathcal{F}^l_r$ from the candidates and add to $\tilde{\mathcal{F}}^l$}
        \State{Update coverage scores of remaining kernels}
    \EndWhile
\EndFor
\end{algorithmic}
\end{algorithm}

\subsection{Complete Pipeline of REPrune}

We introduce REPrune, an efficient pruning pipeline enabling concurrent channel exploration within a CNN during training. This approach is motivated by prior research~\cite{liu2017learning,ye2018rethinking,zhao2019variational}, which leverages the trainable scaling factors ($\gamma$) in Batch Normalization layers to assess channel importance during the training process.
\footnotetext{The number of original in-channels $n_l$ of not pruned filters $\tilde{\mathcal{F}}^l$ is not the same as the out-channel $\lceil (1-s^{l-1})n_l \rceil$ of $\tilde{\mathcal{F}}^{l-1}$. To address this issue, we use a hard pruning method~\cite{he2018soft} that adjusts the $n_l$ in-channels to match $\lceil (1-s^{l-1})n_l \rceil$ of $\tilde{\mathcal{F}}^{l-1}$. }
We define a set of scaling factors across all layers as $\Gamma = \{ \gamma^l_i: (i, l) \in (\mathcal{I}, \mathcal{L}) \}$.
A quantile function, $Q: [0,1] \longmapsto \mathbb{R}$, is applied to $\Gamma$ using a global channel sparsity ratio $\bar{s} \in [0, 1)$. This function determines a threshold $\gamma^*$=$Q(\bar{s};\Gamma)$, which is used to prune non-critical channels:
\begin{equation} \label{global_cutoff}
    Q(\bar{s}; \Gamma) = \text{inf}\{ \gamma^l_i \in \Gamma : F(\gamma^l_i) \geq \bar{s},  (i,l)\in(\mathcal{I}, \mathcal{L}) \},
\end{equation}
where $F(\cdot)$ denotes the cumulative distribution function. The layer-wise channel sparsity $s^l$ can then be derived using this threshold as follows:
\begin{equation} \label{layerwise_channel_sparsity}
s^l=\frac{1}{| \mathcal{I} |}\sum\nolimits_{i\in{\mathcal{I}}}\mathds{1}(\gamma^l_i \leq \gamma^*), \quad  l \in \mathcal{L},
\end{equation}
where the indicator function, $\mathds{1}(\cdot)$, returns 0 if $\gamma^l_j \leq \gamma^*$ is satisfied and 1 otherwise. This allows channels in each layer to be automatically pruned if their corresponding scaling factors are below $\gamma^*$.

In some cases, Eq.\ref{layerwise_channel_sparsity} may result in $s^l$ being equal to one. This indicates that all channels in the $l$-th layer must be removed at specific training iterations, which consequently prohibits agglomerative clustering. We incorporate the \textit{channel regrowth strategy}~\cite{hou2022chex} to tackle this problem. 
This strategy allows for restoring some channels pruned in previous training iterations.
In light of this auxiliary approach, REPrune seamlessly functions without interruption during the training process.

This paper introduces an inexpensive framework to serve REPrune while simultaneously training the CNN model. The following section will demonstrate empirical evaluations of how this proposed method, fully summarized in Alg.~\ref{TotProcess}, surpasses the performance of previous techniques.

\begin{algorithm}[!t]
\caption{Overview of REPrune}\label{TotProcess}
\begin{algorithmic}[1]
\Require{A CNN model $\mathcal{M}$ with convolution filters $\{\mathcal{F}^l: \forall l\in \mathcal{L}\}$; agglomerative clustering algorithm $\mathcal{AC}$; total number of channel pruning epochs $T_{prune}$; pruning interval epoch $\Delta T$; global channel sparsity $\bar{s}$; dataset $\mathcal{D}$ }
\Ensure{A pruned model with filters $\{ \tilde{\mathcal{F}}^l: \forall l\in \mathcal{L} \}$}
\State{Initialize $\mathcal{M}$}
\State{$t \leftarrow 1$ \algorithmiccomment{$t$ denotes epoch}}
\While{$\mathcal{M}$ is \textbf{not} converged}    
    \State{Draw mini-batch samples from $\mathcal{D}$} 
    \State{Perform gradient descent on $\mathcal{M}$}
    \If{$t$ mod $\Delta T$ = 0 \textbf{and} $t \leq T_{prune}$}
        \State{Compute channel sparsity $\{ s^l: \forall l\in \mathcal{L} \}$\algorithmiccomment{Eq.\ref{layerwise_channel_sparsity}}}
        \State{Execute channel selection via REPrune\footnotemark \algorithmiccomment{Alg.\ref{REPrune}}}
        \State{Perform the channel regrowth process}
    \EndIf
    \State{$t \leftarrow t + 1$}
\EndWhile
\end{algorithmic}
\end{algorithm}
\footnotetext{When the event where $s^l=1$ is encountered, indicating the pruning of all channels in $l$-th convolutional layer, the sub-routine of REPrune is exceptionally skipped.}

\begin{table*}[!t]
\begin{minipage}[t]{0.329\linewidth}
    \begin{subtable}[t]{\linewidth}
    \centering
    \fontsize{14.0pt}{14.0pt}\selectfont
    \setlength{\tabcolsep}{0.00001pt}
    \renewcommand{\arraystretch}{1.19}
    \begin{adjustbox}{width=\linewidth}
    \begin{tabular}{lcccc}
    \hline
        \textbf{Method}              & \textbf{PT?}          & \textbf{FLOPs}       & \textbf{Top-1}       & \textbf{Epochs} \\ 
        \hline
        \multicolumn{5}{l}{\textit{\textbf{ResNet-18}}}  \\
        Baseline                             & -           & 1.81G        & 69.4\%     & -    \\
        SFP~\cite{he2018soft}                & \ding{51}   & 1.05G        & 67.1\%     & 200  \\
        FPGM~\cite{he2019filter}             & \ding{51}   & 1.05G        & 68.4\%     & 200  \\
        PFP~\cite{Liebenwein2020Provable}    & \ding{51}   & 1.27G        & 67.4\%     & 270  \\
        SCOP-A~\cite{tang2020scop}           & \ding{51}   & 1.10G        & 69.1\%     & 230  \\
        DMCP~\cite{guo2020dmcp}              & \ding{55}   & 1.04G        & 69.0\%     & 150  \\
        SOSP~\cite{nonnenmacher2022sosp}     & \ding{55}   & 1.20G        & 68.7\%     & 128  \\
        \textbf{REPrune}                     & \ding{55}   &\textbf{1.03G}&\textbf{69.2\%}& \textbf{250} \\  \hline
        \multicolumn{5}{l}{\textit{\textbf{ResNet-34}}} \\
        Baseline                             & -           & 3.6G         & 73.3\%     & -    \\
        GReg-1~\cite{wang2021neural}         & \ding{51}   & 2.7G         & 73.5\%     & 180  \\
        GReg-2~\cite{wang2021neural}         & \ding{51}   & 2.7G         & 73.6\%     & 180  \\
        DTP~\cite{li2023differentiable}      & \ding{55}   & 2.7G         & 74.2\%     & 180  \\
        \textbf{REPrune}                     & \ding{55}& \textbf{2.7G} & \textbf{74.3\%} & \textbf{250} \\ \cdashline{1-5}
        SFP~\cite{he2019filter}              & \ding{51}   & 2.2G         & 71.8\%     & 200  \\
        FPGM~\cite{he2019filter}             & \ding{51}   & 2.2G         & 72.5\%     & 200  \\
        CNN-FCF~\cite{li2019compressing}     & \ding{55}   & 2.1G         & 72.7\%     & 150  \\
        GFS~\cite{ye2020good}                & \ding{51}   & 2.1G         & 72.9\%     & 240  \\
        DMC~\cite{gao2020discrete}           & \ding{51}   & 2.1G         & 72.6\%     & 490  \\
        SCOP-A~\cite{tang2020scop}           & \ding{51}   & 2.1G         & 72.9\%     & 230  \\
        SCOP-B~\cite{tang2020scop}           & \ding{51}   & 2.0G         & 72.6\%     & 230  \\
        NPPM~\cite{gao2021network}           & \ding{51}   & 2.1G         & 73.0\%     & 390  \\
        CHEX~\cite{hou2022chex}              & \ding{55}   & 2.0G         & 73.5\%     & 250  \\
        \textbf{REPrune}                     & \ding{55}   &\textbf{2.0G} &\textbf{73.9\%}& \textbf{250} \\
        \hline
    \end{tabular}
    \end{adjustbox}
    \end{subtable}    
\end{minipage}
\hfill
\begin{minipage}[t]{0.665\linewidth}
    \centering
    \begin{adjustbox}{width=\linewidth}
    \fontsize{14.0pt}{14.0pt}\selectfont
    \setlength{\tabcolsep}{0.00001pt}
    \renewcommand{\arraystretch}{1.2}
    \begin{tabular}{lcccclcccc}
        \hline
        \textbf{Method}                      & \textbf{PT?}& \textbf{FLOPs}& \textbf{Top-1}& \textbf{Epochs}& \textbf{Method}&\textbf{PT?}&\textbf{FLOPs}&\textbf{Top-1}&\textbf{Epochs} \\ 
        \hline
        \multicolumn{5}{l||}{\textit{\textbf{ResNet-50}}}                                             
                                                                                                                        & GroupSparsity~\cite{li2020group}         & \ding{51} & 1.9G & 74.7\%   & - \\
        Baseline                             & -         & 4.1G & 76.2\% & \multicolumn{1}{c||}{-}   
                                                                                                                        & CHIP~\cite{sui2021chip}              & \ding{51} & 2.1G & 76.2\% & 180 \\
        SSS~\cite{huang2018data}             & \ding{55} & 2.3G & 71.8\% & \multicolumn{1}{c||}{100} 
                                                                                                                        & NPPM~\cite{su2021locally}                & \ding{51} & 1.8G & 75.9\%    & 300  \\ 
        SFP~\cite{he2018soft}                & \ding{51} & 2.3G & 74.6\% & \multicolumn{1}{c||}{200}
                                                                                                                        & EKG~\cite{lee2022ensemble}           & \ding{51} & 2.2G & 76.4\% & 300 \\
        MetaPrune~\cite{liu2019metapruning}  & \ding{55} & 3.0G & 76.2\% & \multicolumn{1}{c||}{160}
                                                                                                                        &  HALP~\cite{shen2022structural}       & \ding{51} & 2.0G & 76.5\% & 180 \\
        GBN~\cite{you2019gate}               & \ding{51} & 2.4G & 76.2\% & \multicolumn{1}{c||}{350} 
                                                                                                                        & EKG~\cite{lee2022ensemble}                & \ding{51} & 1.8G & 75.9\%   & 300  \\ 
        FPGM~\cite{he2019filter}             & \ding{51} & 2.3G & 75.5\% & \multicolumn{1}{c||}{200} 
                                                                                                                        & EKG+BYOL\footnotemark[5]~\cite{lee2022ensemble}           & \ding{51} & 1.8G & 76.6\%   & 300 \\ 
        TAS~\cite{dong2019network}           & \ding{55} & 2.3G & 76.2\% & \multicolumn{1}{c||}{240}
                                                                                                                        & DepGraph~\cite{fang2023depgraph}         & \ding{51} & 1.9G & 75.8\%   & -   \\
        GAL~\cite{lin2019towards}            & \ding{51} & 2.3G & 75.0\% & \multicolumn{1}{c||}{570}
                                                                                                                        & \textbf{REPrune}                         & \ding{55} & \textbf{1.8G} & \textbf{77.0\%} & \textbf{250}  \\ \cdashline{6-10}
        PFP~\cite{lin2019towards}            & \ding{51} & 2.3G & 75.2\% & \multicolumn{1}{c||}{270}
                                                                                                                        & CCP~\cite{peng2019ccp}                   & \ding{51} & 1.8G & 75.2\%    & 190 \\
        EagleEye~\cite{li2020eagleeye}       & \ding{51} & 3.0G & 77.1\% & \multicolumn{1}{c||}{240} 
                                                                                                                        & LFPC~\cite{he2020learning}               & \ding{51} & 1.6G & 74.5\%    & 235  \\
        LeGR~\cite{chin2020towards}          & \ding{51} & 2.4G & 75.7\% & \multicolumn{1}{c||}{150} 
                                                                                                                        & Polarization~\cite{zhuang2020neuron}     & \ding{51} & 1.2G & 74.2\%   & 248 \\
        Hrank~\cite{lin2020hrank}            & \ding{51} & 2.3G & 75.0\% & \multicolumn{1}{c||}{570} 
                                                                                                                        & DMCP~\cite{guo2020dmcp} & \ding{55} & 1.1G & 74.1\%     & 150  \\
        HALP~\cite{shen2022structural}       & \ding{51} & 3.1G & 77.2\% & \multicolumn{1}{c||}{180}  
                                                                                                                        & EagleEye~\cite{li2020eagleeye} & \ding{51} & 1.0G & 74.2\%    & 240  \\
        GReg-1~\cite{wang2021neural}         & \ding{51} & 2.7G & 76.3\% & \multicolumn{1}{c||}{180}
                                                                                                                        & ResRep~\cite{ding2021resrep}             & \ding{51} & 1.5G & 75.3\%    & 270  \\
        SOSP~\cite{nonnenmacher2022sosp}     & \ding{55} & 2.4G & 75.8\% & \multicolumn{1}{c||}{128} 
                                                                                                                        & GReg-2~\cite{wang2021neural}             & \ding{51} & 1.3G & 73.9\%    & 180  \\
        \textbf{REPrune}                     & \ding{55} &\textbf{2.3G}& \textbf{77.3\%}& \multicolumn{1}{c||}{\textbf{250}} 
                                                                                                                        & DSNet~\cite{li2021dynamic} & \ding{51} & 1.2G & 74.6\%    & 150  \\
        \cdashline{1-5}
        AdaptDCP~\cite{zhuang2018discrimination} & \ding{51} & 1.9G & 75.2\%   & \multicolumn{1}{c||}{210}  
                                                                                                                        & CHIP~\cite{sui2021chip} & \ding{51} & 1.0G & 73.3\%     & 180  \\
        Taylor-FO~\cite{ding2019centripetal} & \ding{51} & 2.2G & 74.5\% & \multicolumn{1}{c||}{-}   
                                                                                                                        & CafeNet~\cite{su2021locally} & \ding{55} & 1.0G & 75.3\%    & 300  \\ 
        C-SGD~\cite{ding2019centripetal}     & \ding{51} & 2.2G & 74.9\% & \multicolumn{1}{c||}{-}   
                                                                                                                        & HALP+EagleEye~\cite{shen2022structural}  & \ding{51} & 1.2G & 74.5\%    & 180  \\
        MetaPrune~\cite{liu2019metapruning}      & \ding{55} & 2.0G & 75.4\%   & \multicolumn{1}{c||}{160}
                                                                                                                        & SOSP~\cite{nonnenmacher2022sosp} & \ding{55} & 1.1G & 73.3\%    & 128  \\
        SCOP-A~\cite{tang2020scop}           & \ding{51} & 2.2G & 76.0\% & \multicolumn{1}{c||}{230} 
                                                                                                                        & HALP~\cite{shen2022structural} & \ding{51} & 1.0G & 74.3\%    & 180    \\
        DSA~\cite{ning2020dsa}               & \ding{55} & 2.0G & 74.7\% & \multicolumn{1}{c||}{120} 
                                                                                                                        & DTP~\cite{li2023differentiable}          & \ding{55} & 1.7G & 75.5\%    & 180  \\
        EagleEye~\cite{li2020eagleeye}       & \ding{51} & 2.0G & 76.4\% & \multicolumn{1}{c||}{240} 
                                                                                                                        & DTP~\cite{li2023differentiable}          & \ding{55} & 1.3G & 74.2\%    & 180  \\
        SCP~\cite{kang2020operation}             & \ding{55} & 1.9G & 75.3\%   & \multicolumn{1}{c||}{200}
                                                                                                                        &  \textbf{REPrune} & \ding{55} & \textbf{1.0G} & \textbf{75.7\%} & \textbf{250} \\
        \hline
    \end{tabular}
    \end{adjustbox}
\end{minipage}
\caption{Top-1 accuracy comparison between REPrune and prior influential channel pruning methods on the ImageNet validation dataset. `PT?' indicates whether a method necessitates pre-training the original CNN model as part of the train-pruning-finetuning pipeline (\ding{51}) or if it adheres to a concurrent training-pruning paradigm (\ding{55}). `Epochs' represents the entire training time for methods without pre-training CNN; for those requiring pre-training, it denotes the sum of training and fine-tuning time.}
\label{table1}
\end{table*}

\section{Experiment}

This section extensively evaluates REPrune across image recognition and object detection tasks. We delve into case studies investigating the cluster coverage ratio during our proposed MCP optimization. Furthermore, we analyze the impacts of agglomerative clustering with other monotonic distances. We also present REPrune's computational efficiency in the training-pruning time and the image throughput at test time on AI computing devices.

\paragraph{Datasets and models}
We evaluate image recognition on CIFAR-10 and ImageNet~\cite{deng2009imagenet} datasets and object detection on COCO-2017~\cite{lin2014microsoft}. For image recognition, we use various ResNets~\cite{he2016deep}, while for object detection, we employ SSD300~\cite{liu2016ssd}.

\paragraph{Evaluation settings}
This paper evaluates the effectiveness of REPrune using the PyTorch framework, building on the generic training strategy from DeepLearningExample~\cite{shen2022structural}. While evaluations are conducted on NVIDIA RTX A6000 with 8 GPUs, for the CIFAR-10 dataset, we utilize just a single GPU.
Additionally, we employ NVIDIA Jetson TX2 to assess the image throughput of our pruned model with A6000.
Comprehensive details regarding training hyper-parameters and pruning strategies for CNNs are available in the Appendix. This paper computes FLOPs by treating both multiplications and additions as a single operation, consistent with the approach~\cite{he2016deep}.

\subsection{Image Recognition}

\paragraph{Comparison with channel pruning methods} Table~\ref{table1} shows the performance of REPrune when applied to ResNet-18, ResNet-34, and ResNet-50 on the ImageNet dataset.
REPrune demonstrates notable efficiency, with only a slight decrease or increase in accuracy at the smallest FLOPs compared to the baseline. Specifically, the accuracy drop is only 0.2\% and 0.5\% for ResNet-18 and ResNet-50 models, concerning FLOPs of 1.03G and 1.0G, respectively. Moreover, the accuracy of ResNet-34 improves by 0.6\% when its FLOPs reach 2.0G.
Beyond this, REPrune's accuracy not only surpasses the achievements of contemporary influential channel pruning techniques but also exhibits standout results, particularly for the 2.7G FLOPs of ResNet-34 and the 2.3G and 1.8G FLOPs of ResNet-50--all of which display about 1.0\% improvements in accuracy over their baselines.
\footnotetext[5]{BYOL~\cite{grill2020bootstrap} is a self-supervised learning method.}

\begin{table}[!t]
    \begin{adjustbox}{width=\columnwidth,center}
    \setlength{\tabcolsep}{1.0pt}
    \renewcommand{\arraystretch}{1.0}
    \begin{tabular}{lcccccc}
    \hline
    \textbf{Method}                 & \textbf{PT?} & \textbf{\begin{tabular}[c]{@{}c@{}}FLOPs\\ reduction\end{tabular}} & \textbf{\begin{tabular}[c]{@{}c@{}}Baseline\\ Top-1\end{tabular}} & \textbf{\begin{tabular}[c]{@{}c@{}}Pruned\\ Top-1\end{tabular}} & \textbf{\begin{tabular}[c]{@{}c@{}}P-B\\ ($\mathbf{\Delta}$)\end{tabular}} & \textbf{Epochs} \\ \hline
    \multicolumn{7}{l}{\textit{\textbf{ResNet-56 (FLOPs: 127M)}}} \\
    CUP~\cite{Duggal2019CUPCP}      & \ding{55}    & 52.83\%          & 93.67\%       & 93.36\%             & $-$0.31\%      & 360         \\
    LSC~\cite{lee2020filter}\footnotemark[6]        & \ding{51}    & 55.45\%          & 93.39\%       & 93.16\%       & $-$0.23\%      & 1160         \\
    ACP~\cite{chang2022automatic}   & \ding{51}    & 54.42\%          & 93.18\%       & 93.39\%       & $+$0.21\%      & 530         \\ 
    \textbf{REPrune}                & \ding{55}    & \textbf{60.38\%} & \textbf{93.39\%}  & \textbf{93.40\%}   & \textbf{$+$0.01\%} & \textbf{160}                \\ \hline \hline
    \multicolumn{7}{l}{\textit{\textbf{ResNet-18 (FLOPs: 1.81G)}}} \\
    CUP~\cite{Duggal2019CUPCP}      & \ding{55}    & 43.09\%          & 69.87\%       & 67.37\%       & $-$2.50\%      & 180         \\
    ACP~\cite{chang2022automatic}   & \ding{51}    & 34.17\%          & 70.02\%       & 67.82\%       & $-$2.20\%      & 280         \\ 
    \textbf{REPrune}                & \ding{55}    & \textbf{43.09\%}  & \textbf{69.40\%}  &\textbf{69.20\%}& \textbf{$-$0.20\%}  & \textbf{250}     \\ \hline
    \multicolumn{7}{l}{\textit{\textbf{ResNet-50 (FLOPs: 4.1G)}}} \\
    CUP~\cite{Duggal2019CUPCP}      & \ding{55}    & 54.63\%          & 75.87\%       & 74.34\%             & $-$1.47\%      & 180         \\
    ACP~\cite{chang2022automatic}   & \ding{51}    & 46.82\%          & 75.94\%       & 75.53\%       & $-$0.41\%      & 280         \\ 
    \textbf{REPrune}                & \ding{55}    & \textbf{56.09\%}  & \textbf{76.20\%}  &\textbf{77.04\%}& \textbf{$+$0.84\%}  & \textbf{250}     \\ \hline
    \end{tabular}
    \end{adjustbox}
\caption{Top-1 accuracy comparison with prior clustering-based channel pruning methods, using ResNet-56 on CIFAR-10 and ResNet-18, ResNet-50 on ImageNet dataset.}
\label{table3}
\end{table}

\begin{table}[!t]
    \begin{adjustbox}{width=\columnwidth,center}
    \setlength{\tabcolsep}{1.0pt}
    \renewcommand{\arraystretch}{1.0}
    \begin{tabular}{lcccccc}
    \hline
    \textbf{Method}                 & \textbf{PT?} & \textbf{\begin{tabular}[c]{@{}c@{}}FLOPs\\ reduction\end{tabular}} & \textbf{\begin{tabular}[c]{@{}c@{}}Baseline\\ Top-1\end{tabular}} & \textbf{\begin{tabular}[c]{@{}c@{}}Pruned\\ Top-1\end{tabular}} & \textbf{\begin{tabular}[c]{@{}c@{}}P-B\\ ($\mathbf{\Delta}$)\end{tabular}} & \textbf{Epochs} \\ \hline
    \multicolumn{7}{l}{\textit{\textbf{ResNet-56 (FLOPs: 127M)}}} \\
    TMI-GKP~\cite{zhong2022revisit} & \ding{55}    & 43.23\%          & 93.78\%   & 94.00\%      & $+$0.22\%       & 300              \\
    \textbf{REPrune}                & \ding{55}    & \textbf{47.57\%}  & \textbf{93.39\%}  &\textbf{94.00\%}& \textbf{$+$0.61\%}  & \textbf{160}     \\ \hline
    \multicolumn{7}{l}{\textit{\textbf{ResNet-50 (FLOPs: 4.1G)}}} \\
    TMI-GKP~\cite{zhong2022revisit} & \ding{55}    & 33.74\%           & 76.15\%           & 75.53\%        & $-$0.62\%  & 90               \\
    \textbf{REPrune}                & \ding{55}    & \textbf{56.09\%}  & \textbf{76.20\%}  &\textbf{77.04\%}& \textbf{$+$0.84\%}  & \textbf{250}     \\ \hline
    \end{tabular}
    \end{adjustbox}  
\caption{Top-1 accuracy comparison with a recent kernel pruning method, using ResNet-56 and ResNet-50 on each CIFAR-10 and ImageNet dataset.}
\label{table4}
\end{table}


\paragraph{Comparison with clustering-based methods} As shown in Table~\ref{table3}, REPrune stands out with FLOPs reduction rates of 60.38\% for ResNet-56 and 56.09\% for ResNet-50. Each accuracy is compelling, with a slight increase of 0.01\% for ResNet-56 and a distinctive gain of 0.84\% over the baseline for ResNet-50. At these acceleration rates, REPrune's accuracy of 93.40\% (ResNet-56) and 77.04\% (ResNet-50) surpass those achieved by prior clustering-based methods, even at their lower acceleration rates.
\footnotetext[6]{There is no reported accuracy on the ImageNet dataset.}

\paragraph{Comparison with kernel pruning method}
As shown in Table~\ref{table4}, REPrune excels by achieving an FLOPs reduction rate of 56.09\% for ResNet-50 on ImageNet. In this FLOPs gain, REPrune even records a 0.84\% accuracy improvement to the baseline. This stands out when compared to TMI-GKP. On CIFAR-10, REPrune achieves greater FLOPs reduction than TMI-GKP while maintaining the same level of performance.


\begin{table}[!t]
    \begin{adjustbox}{width=\columnwidth,center}
    \setlength{\tabcolsep}{1.0pt}
    \renewcommand{\arraystretch}{1.0}
    \begin{tabular}{lccccccc}
        \hline
        \textbf{Method} & \textbf{\begin{tabular}[c]{@{}c@{}}FLOPs\\ reduction\end{tabular}} & \textbf{mAP} & \textbf{AP$_{50}$} & \textbf{AP$_{75}$} & \textbf{AP$_S$} & \textbf{AP$_M$} & \textbf{AP$_L$} \\ \hline
        \multicolumn{8}{l}{\textit{\textbf{SSD300 object detection}}} \\
        Baseline & 0\% & 25.2 & 42.7 & 25.8 & 7.3 & 27.1 & 40.8 \\
        DMCP~\cite{guo2020dmcp} & 50\% & 24.1 & 41.2 & 24.7 & 6.7 & 25.6 & 39.2 \\
        \textbf{REPrune} & \textbf{50\%} & \textbf{25.0} & \textbf{42.3} & \textbf{25.9} & \textbf{7.4} & \textbf{26.8} & \textbf{40.4} \\
        \hline
    \end{tabular}
    \end{adjustbox}
    
    \caption{Evaluation of SSD300 with ResNet-50 the backbone on COCO-2017. Performance is assessed using bounding box AP. FLOPs reduction refers solely to the backbone.}
    \label{table2}
\end{table}

\subsection{Object Detection}
Table~\ref{table2} presents that, even with a FLOPs reduction of 50\% in the ResNet-50 backbone, REPrune surpasses DMCP by 0.9 in mAP, a minor decrease of 0.2 compared to the baseline.

\subsection{Case Study}

\begin{table}[!t]
    \begin{adjustbox}{width=\columnwidth,center}
    \begin{tabular}{lcccc}
        \hline
        \multicolumn{1}{l|}{\textbf{Linkage method}}    & \textbf{Single}  & \textbf{Complete} & \textbf{Average} & \textbf{Ward} \\ \hline
        \multicolumn{1}{l|}{FLOPs reduction}   &  60.70\%    &  64.04\%    &  63.96\%    & 60.38\%     \\
        \multicolumn{1}{l|}{Top-1 accuracy}   &  92.74\%    &  92.67\%    &  92.87\%    & 93.40\%     \\ \hline
    \end{tabular}
    \end{adjustbox}
    
    \caption{Comparison of linkage methods for ResNet-56 on CIFAR-10, using the same global channel sparsity, $\bar{s}$=0.55.}
    \label{table5}
\end{table}

\paragraph{Impact of other monotonic linkage distance} \label{case_study_1}
Agglomerative clustering is available with other linkage methods, including single, complete, and average, each of which satisfies monotonicity for bottom-up clustering. As shown in Table~\ref{table5}, when Ward's linkage method is replaced with any of these alternatives, accuracy retention appears limited, especially when imposing the same channel sparsity of 55\%.
In the context of the FLOPs reduction with the same channel sparsity, we conjecture that using these alternatives in REPrune tends to favor removing channels from the front layers compared to Ward's method. These layers have more massive FLOPs and are more information-sensitive~\cite{li2017pruning}. While each technique reduces a relatively high number of FLOPs, they may also suffer from maintaining accuracy due to removing these sensitive channels.
\begin{figure}[t!]
    \centering
         \includegraphics[width=1.0\linewidth]{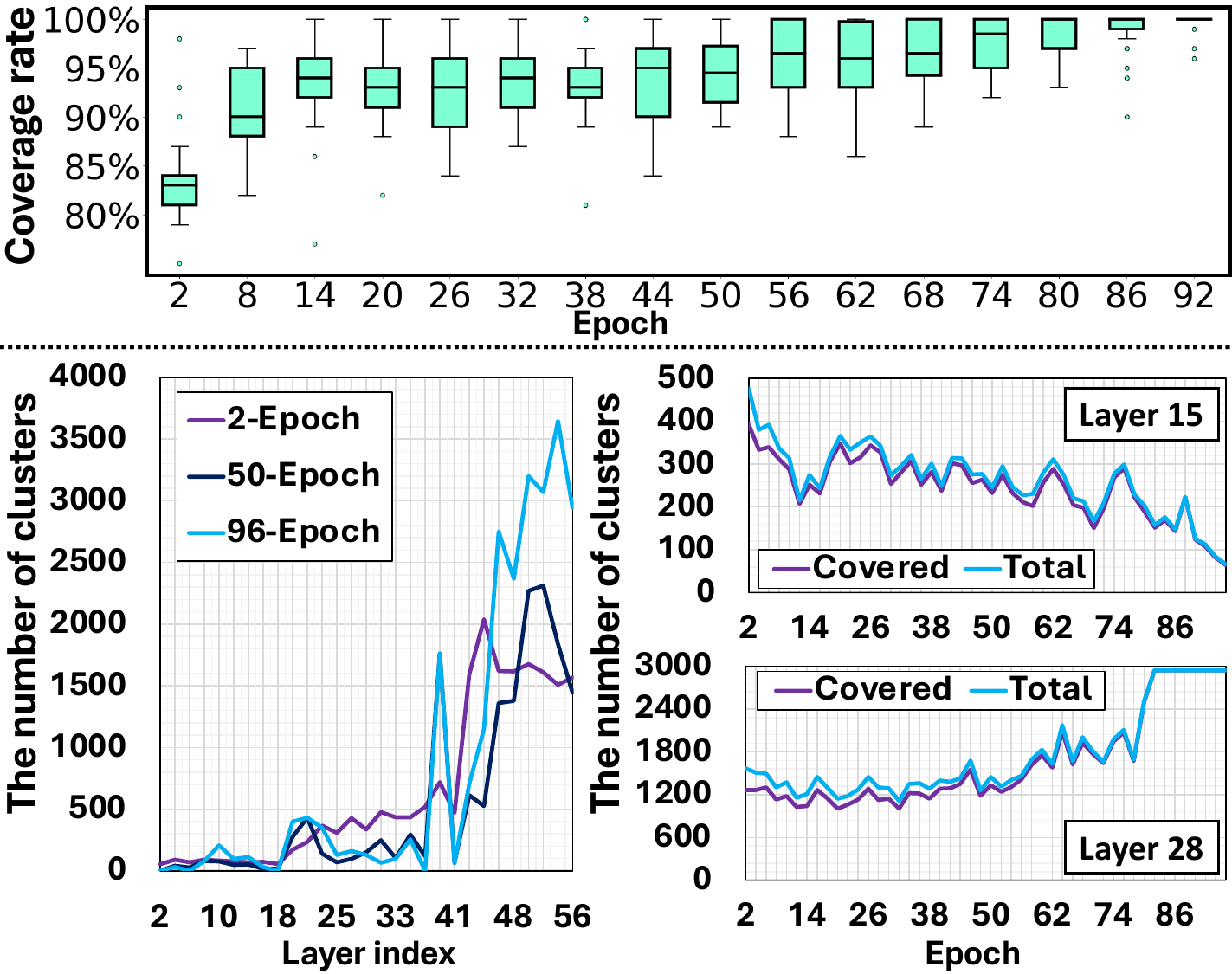}
     \caption{The illustration of coverage rates for ResNet-56 on CIFAR-10 during the optimization of our proposed MCP. Each box contains the coverage rates from all pruned convolutional layers throughout the training epoch.}
     \label{fig_coverage_rate}
\end{figure}

\begin{table}[!t]
    \begin{adjustbox}{width=\columnwidth,center}
    \begin{tabular}{lcccc}
        \hline
        \multicolumn{1}{l|}{\textbf{FLOPs reduction}}    & \textbf{0\%}  & \textbf{43\%} & \textbf{56\%} & \textbf{75\%} \\ \hline
        \multicolumn{1}{l|}{Training time (hours)}   &  43.75    &  40.62    &  38.75    & 35.69     \\ \hline
    \end{tabular}
    \end{adjustbox}
    
    \caption{Training time comparison for ResNet-50 using the REPrune on ImageNet. The `1 hour' is equivalent to training with 8 RTX A6000 GPUs with 16 worker processes.}
    \label{table6}
\end{table}

\begin{figure}[t!]
    \centering
         \includegraphics[width=1.0\linewidth]{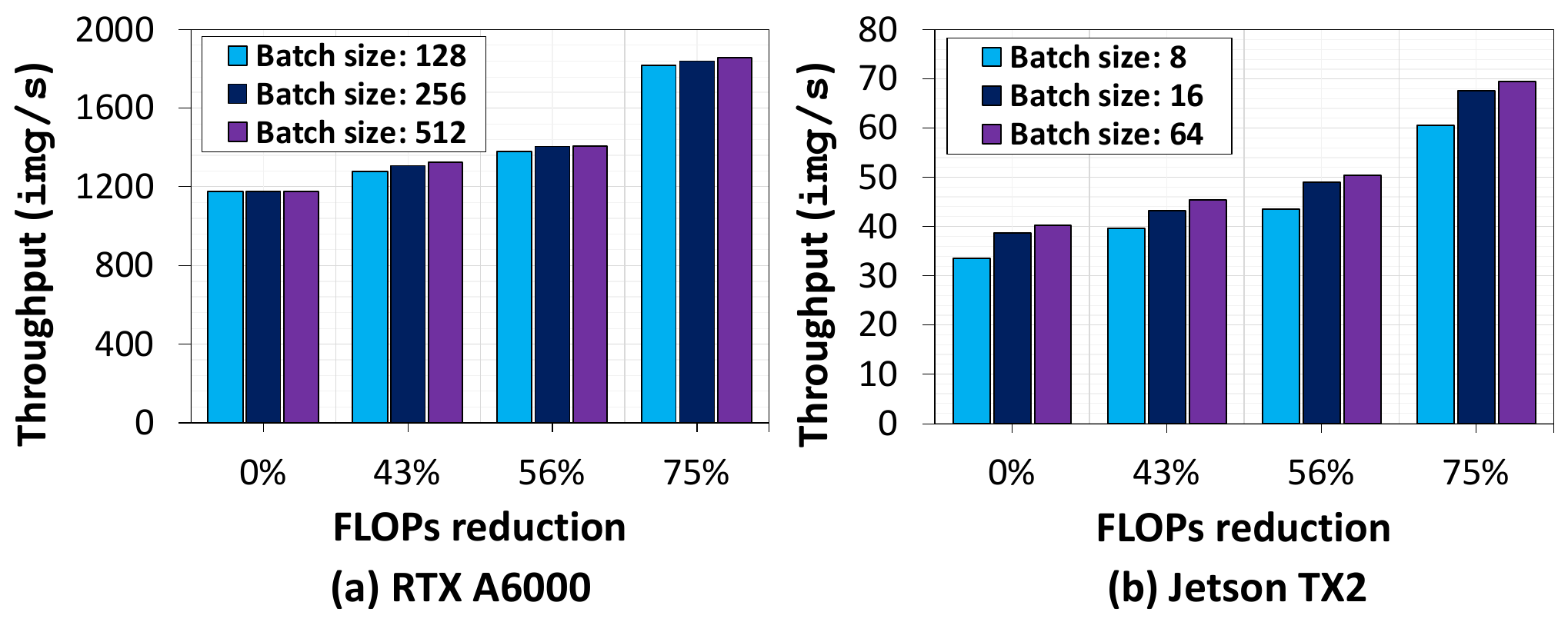}
     \caption{Computing throughput (\texttt{img/s}) on image inference using ResNet-50 on the ImageNet validation dataset.}
     \label{throughput}
\end{figure}

\paragraph{Cluster coverage rate of selected filters}
Fig.~\ref{fig_coverage_rate} illustrates how coverage rates change over time, aggregating selected filter scores relative to the sum of optimal coverage scores.
Initially, clusters from subsequent layers predominate those from the front layers. As learning continues, REPrune progressively focuses on enlarging the channel sparsity of front layers. This dynamic, when coupled with our greedy algorithm's thorough optimization of the MCP across all layers, results in a steady increase in coverage rates, reducing the potential for coverage failures.

\paragraph{Training-pruning time efficiency} Table~\ref{table6} shows the training (forward and backward) time efficiency of REPrune applied to ResNet-50 on the ImageNet dataset. REPrune results in time savings of 3.13, 5.00, and 8.06 hours for FLOPs reductions of 43\%, 56\%, and 75\%, respectively. This indicates that it can reduce training time effectively in practice.

\paragraph{Test-time image throughput}
Fig.~\ref{throughput} demonstrates the image throughput of pruned ResNet-50 models on both a single RTX A6000 and a low-power Jetson TX2 GPU in realistic acceleration.
These models were theoretically reduced by 43\%, 56\%, and 75\% in FLOPs and were executed in various batch sizes. Discrepancies between theoretical and actual performance may arise from I/O delay, frequent context switching, and the number of CUDA and tensor cores.

\section{Conclusion} \label{conclusion_futurework}

Channel pruning techniques suffer from large pruning granularity. 
To overcome this limitation, we introduce REPrune, a new approach that aggressively exploits the similarity between kernels. This method identifies filters that contain representative kernels and maximizes the coverage of kernels in each layer. REPrune takes advantage of both kernel pruning and channel pruning during the training of CNNs, and it can preserve performance even when acceleration rates are high.

\section*{Acknowledgements}
This work was supported by Institute of Information \& communications Technology Planning \& Evaluation (IITP) grant funded by the Korea government (MSIT) (No.2021-0-00456, Development of Ultra-high Speech Quality Technology for remote Multi-speaker Conference System), and by the Korea Institute of Science and Technology (KIST) Institutional Program.

\appendix
\section{Appendix}
\subsection{A.1 Preliminary: Ward's Linkage Distance}
This section introduces defining Ward's linkage distance, denoted by $I(A_c, B_c)$,  for two arbitrary clusters $A_c$ and $B_c$, where each cluster has a finite number of convolutional kernels $\kappa^l_{i,j}$. The linkage distance is given by $I(A_c, B_c)=\Delta(A_c\cup B_c)-\Delta(A_c)-\Delta(B_c)$ where $\Delta(\cdot)$ denotes the Sum of Squared Error (SSE) for a cluster.

Now, when the two clusters are merged, the SSE for the merged cluster, $\Delta(A_c\cup B_c)$, is defined as the sum of squared Euclidean distances between each kernel in the merged cluster and the mean kernel of the merged cluster, $\mathbf{m}_{A_c \cup B_c}$:
\begin{equation} \label{sse_union}
    \Delta(A_c\cup B_c) = \sum_{\kappa^l_{i,j}\in A_c\cup B_c}{\left \| \kappa^l_{i,j} - \mathbf{m}_{A_c \cup B_c} \right \|}_2. \nonumber
\end{equation}
We can break down this SSE into two separate sums, one over kernels in \(A_c\) and one over kernels in \(B_c\):
\begin{align}
    &\sum_{\kappa^l_{i,j}\in A_c\cup B_c}{\left \| \kappa^l_{i,j} - \mathbf{m}_{A_c \cup B_c} \right \|}_2 \nonumber \\
    &= \sum_{\kappa^l_{i,j}\in A_c}{\left \| \kappa^l_{i,j} - \mathbf{m}_{A_c \cup B_c} \right \|}_2 + \sum_{\kappa^l_{i,j}\in B_c}{\left \| \kappa^l_{i,j} - \mathbf{m}_{A_c \cup B_c} \right \|}_2. \nonumber
\end{align}
We further decompose the first term of the SSE using algebraic manipulation:
\begin{align}
    &\sum_{\kappa^l_{i,j}\in A_c}{\left \| \kappa^l_{i,j} - \mathbf{m}_{A_c \cup B_c} \right \|}_2 \nonumber \\
    &= \sum_{\kappa^l_{i,j} \in A_c }\left\| \kappa^l_{i,j} - \left ({\mathbf{m}}_{A_c} - \frac{\left| B_c \right|}{\left| A_c \right| + \left| B_c \right|} ( {\mathbf{m}}_{A_c}- {\mathbf{m}}_{B_c} ) \right ) \right\|_2 \nonumber \\
    &= \sum_{\kappa^l_{i,j} \in A_c }\left\| ( \kappa^l_{i,j} -  {\mathbf{m}}_{A_c} ) + \frac{\left| B_c \right|}{\left| A_c \right| + \left| B_c \right|}( {\mathbf{m}}_{A_c}- {\mathbf{m}}_{B_c} ) \right\|_2 \nonumber \\
    &= \sum_{\kappa^l_{i,j} \in A_c }^{} \left\| \kappa^l_{i,j} - {\mathbf{m}}_{A_c} \right\|_2 \nonumber \\
    &+ 2  \sum_{\kappa^l_{i,j} \in A_c } \| \kappa^l_{i,j} - {\mathbf{m}}_{A_c} \| \cdot\ \frac{\left| B_c \right|}{\left| A_c \right| + \left| B_c \right|} \| {\mathbf{m}}_{A_c} - {\mathbf{m}}_{B_c} \| \nonumber \\
    &+ \left| A_c \right|  \left\| \frac{\left| B_c \right|}{\left| A_c \right| + \left| B_c \right|}\left ( {\mathbf{m}}_{A_c} - {\mathbf{m}}_{B_c} \right ) \right\|_2, \nonumber
\end{align}
The term, $\sum_{\kappa^l_{i,j} \in A_c } \| \kappa^l_{i,j} - {\mathbf{m}}_{A_c} \|$, is zero. Hence, the first term of SSE is as follows: 
\begin{align} \label{first_term}
    &\therefore \sum_{\kappa^l_{i,j}\in A_c}{\left \| \kappa^l_{i,j} - \mathbf{m}_{A_c \cup B_c} \right \|}_2 \nonumber \\ 
    &=\sum_{\kappa^l_{i,j} \in A_c } \left\| \kappa^l_{i,j} - {\mathbf{m}}_{A_c} \right\|_2 +  \frac{{\left| A_c \right| }{\left| B_c \right| }^{2}}{\left ( \left| A_c \right|  +\left| B_c \right| \right )^{2}}\left\|  {\mathbf{m}}_{A_c} - {\mathbf{m}}_{B_c} \right\|_2.
\end{align}
Here, $\mathbf{m}_{A_c}$ and $\mathbf{m}_{B_c}$ denote the mean kernels of clusters $A_c$ and $B_c$, respectively.
We can derive the second term of the SSE in a similar manner:
\begin{align} \label{second_term}
    &\therefore \sum_{\kappa^l_{i,j}\in B_c}{\left \| \kappa^l_{i,j} - \mathbf{m}_{A_c \cup B_c} \right \|}_2 \nonumber \\ 
    &=\sum_{\kappa^l_{i,j} \in B_c } \left\| \kappa^l_{i,j} - {\mathbf{m}}_{B_c} \right\|_2 +  \frac{{\left| A_c \right| }^{2}{\left| B_c \right| }}{\left ( \left| A_c \right|  +\left| B_c \right| \right )^{2}}\left\|  {\mathbf{m}}_{B_c} - {\mathbf{m}}_{A_c} \right\|_2.
\end{align}
By Eq.~\ref{first_term} and Eq.~\ref{second_term}, we can rewrite the SSE for the merged cluster:
\begin{align}
    \sum_{\kappa^l_{i,j}\in A_c\cup B_c}{\left \| \kappa^l_{i,j} - \mathbf{m}_{A_c \cup B_c} \right \|}_2 = \nonumber \\
    & \hspace{-4.5cm} \sum_{\kappa^l_{i,j} \in A_c } \left\| \kappa^l_{i,j} - {\mathbf{m}}_{A_c} \right\|_2 + \sum_{\kappa^l_{i,j} \in B_c } \left\| \kappa^l_{i,j} - {\mathbf{m}}_{B_c} \right\|_2  \nonumber \\
    & \hspace{-4.5cm} + \frac{{\left| A_c \right|}{\left| B_c \right|}}{\left| A_c \right| + \left| B_c \right|} \left\| {\mathbf{m}}_{A} - {\mathbf{m}}_{B} \right\|_2.
\end{align}
Finally, we have the last term as Ward's Linkage Distance, $I(A_c, B_c)$:
\begin{equation}
\begin{split}
\therefore I(A_c, B_c) & =\frac{\left | A_c \right | \left | B_c \right |}{\left | A_c \right | + \left |  B_c \right |} {\left \| \mathbf{m}_{A_c} - \mathbf{m}_{B_c} \right \|}_2.
\end{split}
\end{equation}

\subsection{A.2 Coverage Rate Formula}
To quantify the degree of cluster coverage rates after each training epoch, as illustrated in Figure~\ref{fig_coverage_rate}, we introduce a metric from Eq.~\ref{mcp}, denoted as $\mathcal{R}^l_{cov}$ for the $l$-th layer:
\begin{equation}
\mathcal{R}^l_{cov}= \sum\nolimits_{\mathcal{F}^l_r\in\tilde{\mathcal{F}}^l}\sum\nolimits_{\kappa^l_{r,j}\in\mathcal{F}^l_r}\sum\nolimits_{j\in\mathcal{J}}S(\kappa^l_{r,j}) / |U^l|.
\end{equation}
This metric calculates the total coverage score $S(\cdot)$ for selected kernels relative to the maximum possible coverage score $|U^l|$ across all layers $L$ in a CNN model.

\subsection{A.3 Training Settings}
\paragraph{Datasets} ImageNet dataset consists of 1.27M training images and 50k validation images over 1K classes. The COCO-2017 dataset is divided into train2017, 118k images, and val2017, which contains 5k images with 80 object categories. The COCO-2014 provides 83k training images and 41k validation images.

\paragraph{Image recognition} We train ResNet models on the ImageNet dataset using NVIDIA training strategies~\cite{shen2022structural}, \texttt{DeepLearningExamples}, leveraging mixed precision.
Unless otherwise specified, we use default parameters $T_{prune}=180$ and $\Delta T=2$ epochs. Here, $T_{prune}$ indicates the epoch at which channel pruning and regrowth are scheduled, and $\Delta T$ determines the interval to perform the pruning of channels. For data augmentation, we solely adopt the strategies~\cite{he2016deep}, which include random resized cropping to 224$\times$224, random horizontal flipping, and normalization. All models are trained with the same seed.
For training ResNet and MobileNetV2 on CIFAR-10, we follow the strategies used for ImageNet but adjust the weight decay to 5e-4 and the batch size to 256. We train the model for 160 epochs without mixed precision and set $T_{prune}=96$.

\paragraph{Object detection} We set the input size to 300$\times$300 on COCO-2017 training and validation splits. We use stochastic gradient descent (SGD) with a momentum of 0.9, a batch size of 64, and a weight decay of 5e-4. The model is trained over 240k iterations with an initial learning rate of 2.6e-3, which is reduced by a factor of ten at the 160k and 200k iterations. We employ a ResNet-50 from \texttt{torchvision} as the backbone of the SSD300 model, which has been pre-trained on the ImageNet dataset.

\paragraph{Instance segmentation} We follow the common practice as \texttt{DeepLearningExamples} to train Mask R-CNN with the REPrune using COCO-2014 training and validation splits. We train with a batch size of 32 for 160K iterations. We adopt SGD with a momentum of 0.9 and a weight decay of 1e-4. The initial learning rate is set to 0.04, which is decreased by 10 at the 100k and 140k iterations. We use the ResNet-50-FPN model as the backbone.

\paragraph{Training and test environment}
Table~\ref{table7} introduces the hardware resources used in this paper for training and test time. In this paper, we perform distributed parallel training utilizing the resources of the RTX A6000. In the training of ResNet for image recognition, we use a total of 8 GPUs, issuing 128 batches to each, and for object detection, we use a total of 2 GPUs, issuing 32 batches to each duplicated model. The number of worker processes to load image samples is 2 per GPU for image recognition and 8 per GPU for object detection. This paper conducts experiments on a single A6000 GPU and a Jetson TX2 for real-time inference tests. For measuring image throughput on the Jetson TX2, only the four cores of the ARM Cortex-A57 are utilized.

\begin{table}[t!]
\begin{adjustbox}{width=\linewidth,center}
    \setlength{\tabcolsep}{1.2pt}
    \renewcommand{\arraystretch}{1.0}
        \begin{tabular}{l|c|c}
        \hline
        \textbf{Name} & \textbf{RTX A6000}                 & \textbf{Jetson TX2}                                \\ \hline \hline
        CPU  & \begin{tabular}[c]{@{}c@{}} 16$\times$ Intel Xeon \\ Gold 5222 @ 3.8GHz \end{tabular}  & \begin{tabular}[c]{@{}c@{}} 4$\times$ ARM Cotex-A57\\ + 2$\times$ NVIDIA Denver2\end{tabular}      \\ \hline
        RAM  & 512 GB                   & \begin{tabular}[c]{@{}c@{}} 8 GB \\ (shared with GPU)\end{tabular}                   \\ \hline
        GPU  & \begin{tabular}[c]{@{}c@{}} 8 NVIDIA RTX A6000, \\ 48 GB \end{tabular} & \begin{tabular}[c]{@{}c@{}} Tegra X2 (Pascal), \\ 8 GB (shared with RAM) \end{tabular} \\ \hline
        \end{tabular}
    \end{adjustbox}
    \caption{Details of the training and test platform.}
    \label{table7}
\end{table}

\begin{table}[t!]
    \begin{adjustbox}{width=\columnwidth,center}
    \setlength{\tabcolsep}{1.0pt}
    \renewcommand{\arraystretch}{1.0}
    \begin{tabular}{lccccc}
    \hline
    \textbf{Method}            & \textbf{PT?} & \textbf{\begin{tabular}[c]{@{}c@{}}FLOPs\\ reduction\end{tabular}} & \textbf{\begin{tabular}[c]{@{}c@{}}Baseline\\ Top-1\end{tabular}} & \textbf{\begin{tabular}[c]{@{}c@{}}Pruned\\ Top-1\end{tabular}} &  \textbf{Epochs} \\ \hline
    \multicolumn{6}{l}{\textit{\textbf{MobileNetV2 (FLOPs: 296M)}}} \\
    DCP~\cite{zhuang2018discrimination} & \ding{51}    & 26.40\%   &  94.47\%   & 94.02\%    & 400 \\
    MDP~\cite{guo2020mdp}               & \ding{51}    & 28.71\%   &  95.02\%   & 95.14\%    & -   \\
    DMC~\cite{gao2020discrete}          & \ding{51}    & 40.00\%   &  94.23\%   & 94.49\%    & 360 \\
    SCOP~\cite{guo2021gdp}              & \ding{51}    & 40.30\%   &  94.48\%   & 94.24\%    & 400 \\
    \textbf{REPrune}                    & \ding{55}    & \textbf{43.24\%}   &  94.94\%         & \textbf{95.14\%}  & \textbf{160} \\ \cdashline{1-6}
    AAP~\cite{zhao2023aap}              & \ding{55}    & 55.67\%   &  94.46\%   & 94.74\%    & 300 \\
    \textbf{REPrune}                    & \ding{55}    & \textbf{56.08\%}   &  94.94\%         & \textbf{94.74\%}  & \textbf{160} \\ \hline
    \end{tabular}
    \end{adjustbox}
\caption{Top-1 accuracy comparison with the recent channel pruning methods using MobileNetV2 on CIFAR-10 dataset.}
\label{table8}
\end{table}

\subsection{A.4 Supplementary Experiment}
\paragraph{REPrune on MobileNetV2}
The inherent computational efficiency of MobileNetV2, achieved via depthwise separable convolutions, makes channel pruning particularly challenging.
However, as shown in Table~\ref{table8}, REPrune surpasses previous channel pruning techniques, improving accuracy in FLOPs reduction scenarios of 43.24\% and 56.08\%, despite these challenges. In these scenarios, REPrune achieves higher accuracies of 95.14\% and 94.74\% while reducing more FLOPs than earlier methods. Remarkably, REPrune also uses solely 160 epochs, which reduce training time by 46\% to 60\% compared to traditional methods, presenting greater efficiency in both speed-up and accuracy.

\paragraph{Impact of the channel pruning epoch $T_{prune}$}
Table~\ref{table9} presents the accuracy of ResNet-18 trained on the ImageNet dataset, taking into account the changes in the time ($T_{prune}$) dedicated to concurrent pruning and channel regrowth within the overall training time. $T_{prune}$ is determined to fall within a range of 50\% (75) to 100\% (250) of the total training epochs. The results reveal that a $T_{prune}$ of 180, equivalent to 72\% of the total training time, achieves the highest accuracy for the pruned model.

\begin{table}[t!]
    \begin{adjustbox}{width=\linewidth,center}
            \begin{tabular}{l|cccc}
            \hline
            \textbf{$T_{prune}$} & \textbf{75}& \textbf{125} & \textbf{180} & \textbf{250} \\ \hline
            Top-1 accuracy       & 68.64\%     &  68.93\%      & \textbf{69.20\%}       & 68.72\%       \\  \hline
            \end{tabular}
    \end{adjustbox}
    \caption{$T_{prune}$ shifts in training ResNet-18 on ImageNet.}
    \label{table9}
\end{table}

\begin{table}[t!]
    \begin{adjustbox}{width=\columnwidth,center}
    \setlength{\tabcolsep}{1.0pt}
    \renewcommand{\arraystretch}{1.0}
    \begin{tabular}{lcccccc}
    \hline
    \textbf{Method}                 & \textbf{PT?} & \textbf{\begin{tabular}[c]{@{}c@{}}FLOPs\\ reduction\end{tabular}} & \textbf{\begin{tabular}[c]{@{}c@{}}Baseline\\ Top-1\end{tabular}} & \textbf{\begin{tabular}[c]{@{}c@{}}Pruned\\ Top-1\end{tabular}} & \textbf{\begin{tabular}[c]{@{}c@{}}P-B\\ ($\mathbf{\Delta}$)\end{tabular}} & \textbf{Epochs} \\ \hline
    \multicolumn{7}{l}{\textit{\textbf{ResNet-56 (FLOPs: 127M)}}} \\
    Train-prune-finetune & \ding{51}    & 56.96\% & 93.39\%  & 93.02\%   & $-$0.37\%       & 160+160              \\
    Original (\textbf{Proposed})  & \ding{55}    & \textbf{60.38\%} & \textbf{93.39\%}  & \textbf{93.40\%}   & \textbf{$+$0.01\%} & \textbf{160}                \\ \hline
    \end{tabular}
    \end{adjustbox}
\caption{Comparison of Top-1 accuracy using proposed REPrune versus REPrune based on a traditional train-prune-finetune method with ResNet-56 on CIFAR-10. For the train-prune-finetune method, `Epochs' denotes the total number of epochs, including training and finetuning. We set a global channel sparsity $\bar{s}$ to 0.55 in both cases.}
\label{table10}
\end{table}

\paragraph{Converting train-prune-finetune approach}
We change REPrune, initially designed as a progressive training-pruning strategy, to fit the three-stage pruning approach known as the train-prune-finetune pipeline. We then assessed the effectiveness and efficiency of our selection criterion within this pipeline. Table~\ref{table10} demonstrates that our proposed framework offers three significant advantages over adapting the traditional approach. First, the original REPrune reduces FLOPs by 60.38\% at the same global channel sparsity of 0.55 ($\bar{s}$), leading to a 3.42\% reduction in computation even when the channel count is the same. Second, it achieves a higher accuracy of 93.40\%, indicating better accuracy retention that slightly surpasses the traditional method by 0.38\%. Third, our approach takes half the time to achieve this level of pruning compared to traditional methods, highlighting its time-saving benefits.

\begin{table}[t!]
    \begin{adjustbox}{width=\columnwidth,center}
    \begin{tabular}{lcc||lcc}
        \hline
        \textbf{Random} & \textbf{FLOPs} & \textbf{Top-1} & \textbf{Selection} & \textbf{FLOPs} & \textbf{Top-1} \\ \hline
        1-seed & 1.09G & 69.48\% & Random  & 1.09G & 69.48\%  \\
        2-seed & 1.09G & 69.44\% & Max. $\ell_2$-norm & 1.09G & 69.46\%  \\
        3-seed & 1.08G & 69.52\% & Min. $\ell_2$-norm & 1.10G & 69.57\%  \\
        \hline
    \end{tabular}
    \end{adjustbox}
    \caption{Evaluation using ResNet-18 on ImageNet. The results use the same global channel sparsity $\bar{s}=0.6$}
    \label{table11}
\end{table}

\paragraph{Impact on randomness of the filter selection}
What we want to address is that candidate filters may share a similar mixture from representative kernels. However, their real values can indeed differ.
To investigate the impacts of this variance, we explored how random selection influences FLOPs and Top-1 accuracy by conducting experiments with three unique seeds and two supplementary selection criteria.
Table~\ref{table11} demonstrates that pruning outcomes can vary slightly influenced by variations in random seeds and the selection of candidates; it implies that randomness is still useful, but better selection may exist.

\begin{table}[t!]
    \begin{adjustbox}{width=\columnwidth,center}
    \setlength{\tabcolsep}{3.5pt}
    \renewcommand{\arraystretch}{1.0}
    \begin{tabular}{lccccccc}
        \hline
        \textbf{Method} & \textbf{\begin{tabular}[c]{@{}c@{}}FLOPs\\ reduction\end{tabular}} & \textbf{mAP} & \textbf{AP$_{50}$} & \textbf{AP$_{75}$} & \textbf{AP$_S$} & \textbf{AP$_M$} & \textbf{AP$_L$} \\ \hline
        \multicolumn{8}{l}{\textit{\textbf{Mask R-CNN object detection}}} \\
        Baseline & - & 37.3 & 59.0 & 40.2 & 21.9 & 40.9 & 48.1 \\
        \textbf{REPrune} & \textbf{30\%} & \textbf{36.5} & \textbf{57.5} & \textbf{39.4} & \textbf{20.5} & \textbf{38.8} & \textbf{48.6} \\
        \hline
        \multicolumn{8}{l}{\textit{\textbf{Mask R-CNN instance segmentation}}} \\
        Baseline & - & 34.2 & 55.9 & 36.2 & 15.8 & 36.9 & 50.1 \\
        \textbf{REPrune} & \textbf{30\%} & \textbf{33.8} & \textbf{54.6} & \textbf{35.9} & \textbf{15.1} & \textbf{35.8} & \textbf{50.4} \\
        \hline
    \end{tabular}
    \end{adjustbox}
    \caption{Evaluation of Mask R-CNN with ResNet-50-FPN backbone on COCO-2014. Higher AP values denote better performance. FLOPs reduction refers only to the backbone.}
    \label{table12}
\end{table}

\subsection{A.5 Instance Segmentation}
We evaluate REPrune to the Mask R-CNN, leveraging the COCO-2014 dataset. Mask R-CNN requires the concurrent optimization of three distinct loss functions: $\ell_{cls}$, $\ell_{box}$, and $\ell_{mask}$. This multi-task training shows a challenging scenario for pruning the model's backbone due to its generalization.
Hence, we targeted a 30\% reduction in FLOPs for the backbone and examined the results based on a key metric: the Average Precision (AP) for object detection and instance segmentation.
According to Table~\ref{table12}, REPrune demonstrates a minimal impact on performance, with a decrease of 0.8 mAP in object detection and 0.4 mAP in instance segmentation. The results still present the efficacy of REPrune in accelerating a more demanding computer vision task.

\subsection{A.6 Pruning Details}

\begin{figure}[t!]
    \centering
        \includegraphics[width=1\linewidth]{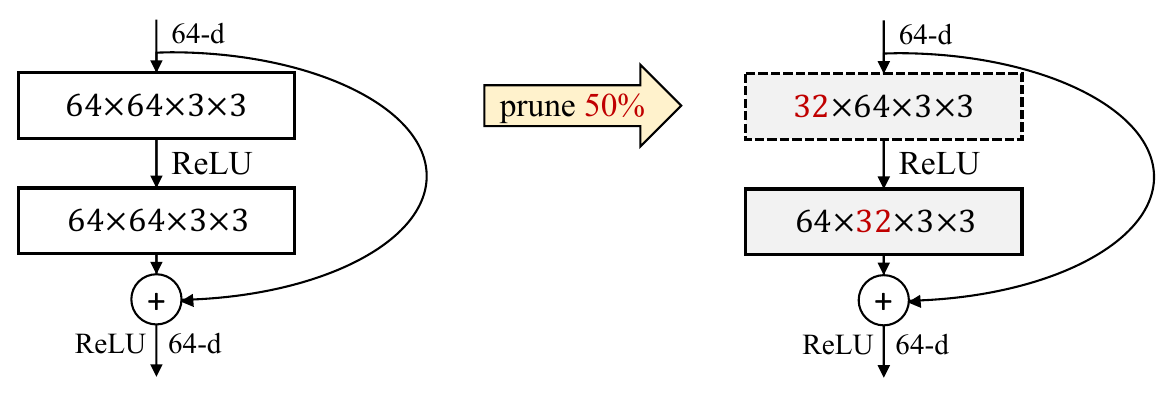}
    \caption{Pruning strategy for the basic residual block, where only the $3\times 3$ convolutional layer marked with dotted lines is pruned by 50\% (assuming $s^l$ is 0.5), leaving the output channel dimension unchanged.}
    \label{basic_pruning}
    \end{figure}

\paragraph{ResNet-18 and ResNet-34} Both ResNet models are composed of multiple basic residual blocks. Figure~\ref{basic_pruning} shows that the basic residual block involves adding the input feature map, which passes through the skip connection, to the output feature map of the second 3x3 convolutional layer within the block. If both of the initial two convolutional layers are pruned in the residual block, there arises a problem with performing the addition operation due to a mismatch in the number of channels between the output feature map of the second $3\times 3$ convolution and the input feature map of the residual block. To circumvent this issue, we apply REPrune exclusively to the first $3\times 3$ convolutional layer, as illustrated in Figure~\ref{basic_pruning}. This strategy is widely accepted and supported by previous works~\cite{he2018soft,he2019filter,huang2018data,luo2017thinet}.

\begin{figure}[t!]
    \centering
        \includegraphics[width=1\linewidth]{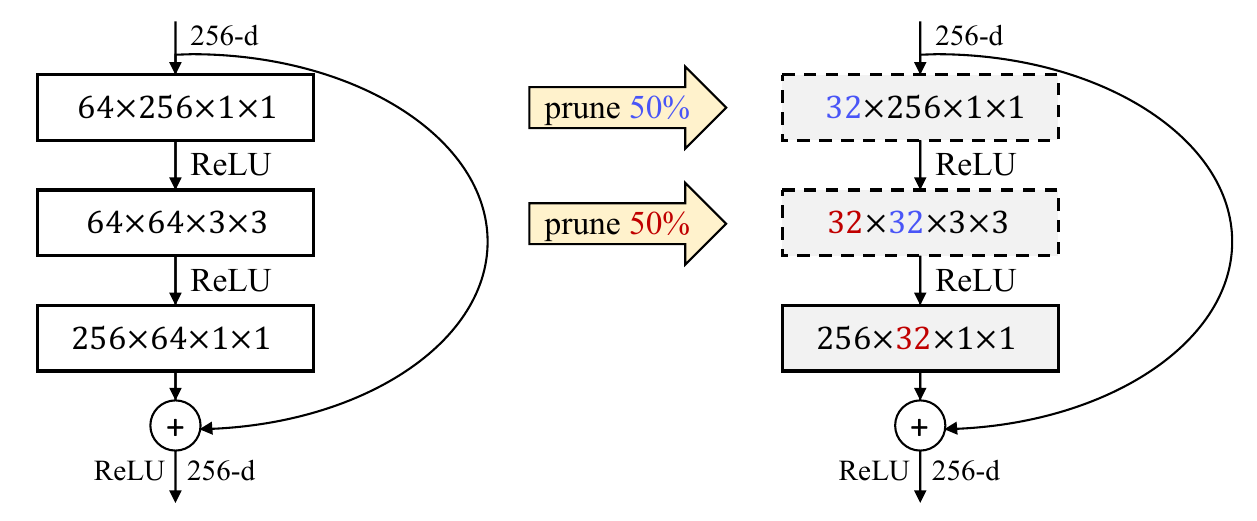}
    \caption{Pruning strategy for the bottleneck block: the first two parts ($1\times 1$ and $3\times 3$) of the three convolutional layers, indicated by dotted lines, are pruned by 50\% (assuming each $s^l$ is 0.5). The third $1\times 1$ layer is not pruned, ensuring the output channel dimensions of both the skip-connection and the last $1\times 1$ convolutional layer remain unchanged.}
    \label{bottleneck_pruning}
    \end{figure}

\paragraph{ResNet-50} This model is built with multiple bottleneck blocks. Figure~\ref{bottleneck_pruning} illustrates that the bottleneck block also adds the input feature map passing through the skip-connection and the output feature map of the last $1\times 1$ convolution layer within the block.
Along with the basic block, if all the convolutional layers are pruned in a bottleneck block, the problem of not performing the addition occurs because the output feature map of the third $1\times 1$ convolution does not match the number of channels with the input feature map of the bottleneck block.
In this paper, we perform REPrune only on the first two convolutional layers.


\fontsize{9.0pt}{10.0pt} \selectfont
\bibliography{aaai24}

\end{document}